\documentclass[10pt,twocolumn,letterpaper]{article}
\pdfoutput=1
\usepackage{iccv}
\usepackage{times}
\usepackage{epsfig}
\usepackage{graphicx}
\usepackage{amsmath}
\usepackage{amssymb}

\usepackage{lipsum}
\usepackage{multirow}
\usepackage{booktabs}

\newcommand{\mymodel}{Reflective Decoding Network\xspace}

\iccvfinalcopy

\def\iccvPaperID{3813} 

\ificcvfinal\fi

\begin{document}
	
	\title{Reflective Decoding Network for Image Captioning}
	
\author{
	Lei Ke$^1$,\hspace{1.0cm}Wenjie Pei$^2$,\hspace{1.0cm}Ruiyu Li$^2$,\hspace{1.0cm}Xiaoyong Shen$^{2}$,\hspace{1.0cm}Yu-Wing Tai$^2$
	\vspace{0.1cm}\\
	$^1$The Hong Kong University of Science and Technology,  $^2$Tencent
	\\
	\texttt{\footnotesize keleiwhu@gmail.com,wenjiecoder@outlook.com,\{royryli,dylanshen,yuwingtai\}@tencent.com }}

	\maketitle
	\thispagestyle{empty}

\begin{abstract}	
State-of-the-art image captioning methods mostly focus on improving visual features, less attention has been paid to utilizing the inherent properties of language to boost captioning performance. 
In this paper, we show that vocabulary coherence between words and syntactic paradigm of sentences are also important to generate high-quality image caption.
Following the conventional encoder-decoder framework, we propose the Reflective Decoding Network (RDN) for image captioning, which enhances both the long-sequence dependency and position perception of words in a caption decoder.
Our model learns to collaboratively attend on both visual and textual features and meanwhile perceive each word's relative position in the sentence to maximize the information delivered in the generated caption.
We evaluate the effectiveness of our RDN on the COCO image captioning datasets and achieve superior performance over the previous methods.
Further experiments reveal that our approach is particularly advantageous for hard cases with complex scenes to describe by captions.


\vspace{-0.2in}
\end{abstract}

\vspace{-2mm}
\section{Introduction}
The goal of image captioning is to automatically generate fluent and informative language description of an image for human understanding.
As an interdisciplinary task connecting Computer Vision and Nature Language Processing, it explores towards
the cutting edge techniques of scene understanding \cite{li2009towards} and it is drawing increasing interests in recent years.
\footnotetext[1]{This work was done while Lei Ke was an intern at Tencent.}

To build a top captioning system, there are two crucial requirements.
First, the captioning model needs to distill representative and meaningful visual representation from an image. 
Thanks to the success in image classification \cite{krizhevsky2012imagenet} and object recognition \cite{he2014spatial,ren2015faster}, recent methods \cite{anderson2017bottom,lu2017knowing,xu2015showattendtell,yang2016review} have shown significant advancements which mostly benefited from the improved quality of extracted visual features.
Second, and the relatively neglected requirement, is to make the generated captions coherent and intelligent. 
Similar to the human language system, it needs to inference and reason during the generation process based on what has been generated and watched.
Typically, this process is achieved by RNN (specifically, LSTM~\cite{hochreiter1997long}) in storing the sequential information during caption decoding.

\begin{figure}[t]
	\centering
	\includegraphics[width=1.0\linewidth]{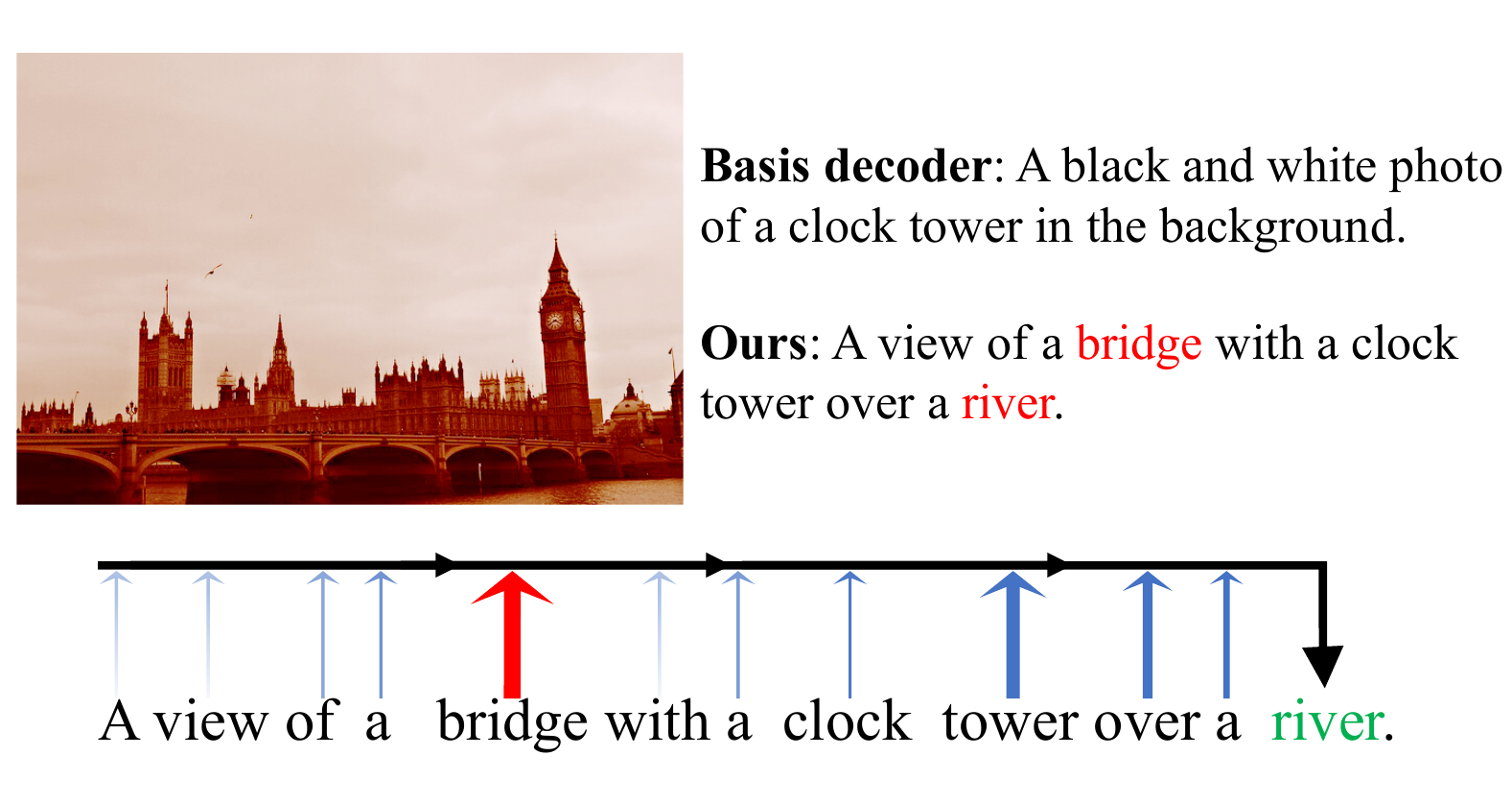}
	\caption{\textbf{Top}: Example captions generated by the basis decoder (using traditional LSTM) and our Reflective Decoding Network model. \textbf{Bottom}: The reflective attention weight distribution over the past generated hidden states is shown when predicting the word `river'. The thicker line indicates a relatively larger weight and the red line means the largest contribution to the prediction.}
	\label{fig:example}
	\vspace{-0.2in}
\end{figure}

The traditional LSTM model, however, tends to focus more on the relatively closer vocabulary while neglecting the farther one.
For example, in Figure~\ref{fig:example}, the word `bridge' has an important hint on predicting the word `river' (which is neglected by the basis decoder), but the two words are separated by 6 words.
Current mainstream caption decoder is weak in handling this kind of long-term dependency in sequential sentence, especially when the visual content of an image is complex and hard to describe, which usually leads to a general and less accurate caption description.

In this paper, we propose the \mymodel(RDN) for image captioning, which mitigates the drawback of traditional caption decoder by enhancing its long sequential modeling ability.
Different from previous methods which boost captioning performance by improving the visual attention mechanism~\cite{anderson2017bottom,lu2017knowing,xu2015showattendtell}, or by improving the encoder to supply more meaningful intermediate representation for the decoder~\cite{jiang2018recurrent,yang2016review,yao2018exploring,you2016image}, our RDN focuses directly on the target decoding side and jointly apply attention mechanism in both visual and textual domain.
Besides, we propose to model the positional information of each word within a caption in a supervised way to capture the syntactic structure of natural language.
Another advantage in RDN is to visualize how the model inferences and makes word prediction based on the generated words.
For instance, our RDN successfully decodes the word `river' in Figure~\ref{fig:example} by referring to the previously generated words, especially the most relevant word `bridge'.

The main contributions of this paper are four folds:
\vspace{-0.05in}
\begin{itemize}
	\item We propose the RDN that effectively enhances the long sequential modeling ability of the traditional caption decoder for generating high-quality image captions.
	\vspace{-0.07in}
	\item By considering long-term textual attention, we explicitly explore the coherence between words and visualize the word decision making process in text domain to show how we can interpret the principle and result of the framework from a novel perspective. 
	\vspace{-0.07in}
	\item We design a novel positional module to enable our RDN to perceive the relative position of each word in the whole caption and thereby better comprehend the syntactic paradigm of natural language.
	\vspace{-0.07in}
    \item Our RDN achieves state-of-the-art performance on COCO captioning dataset and is particularly superior over existing methods in hard cases with complex scenes to describe by captions.
\end{itemize}
\vspace{-0.05in}
\vspace{-2mm}

\section{Related Work}
\begin{figure*}[t]
	\centering
	\includegraphics[width=0.8\linewidth]{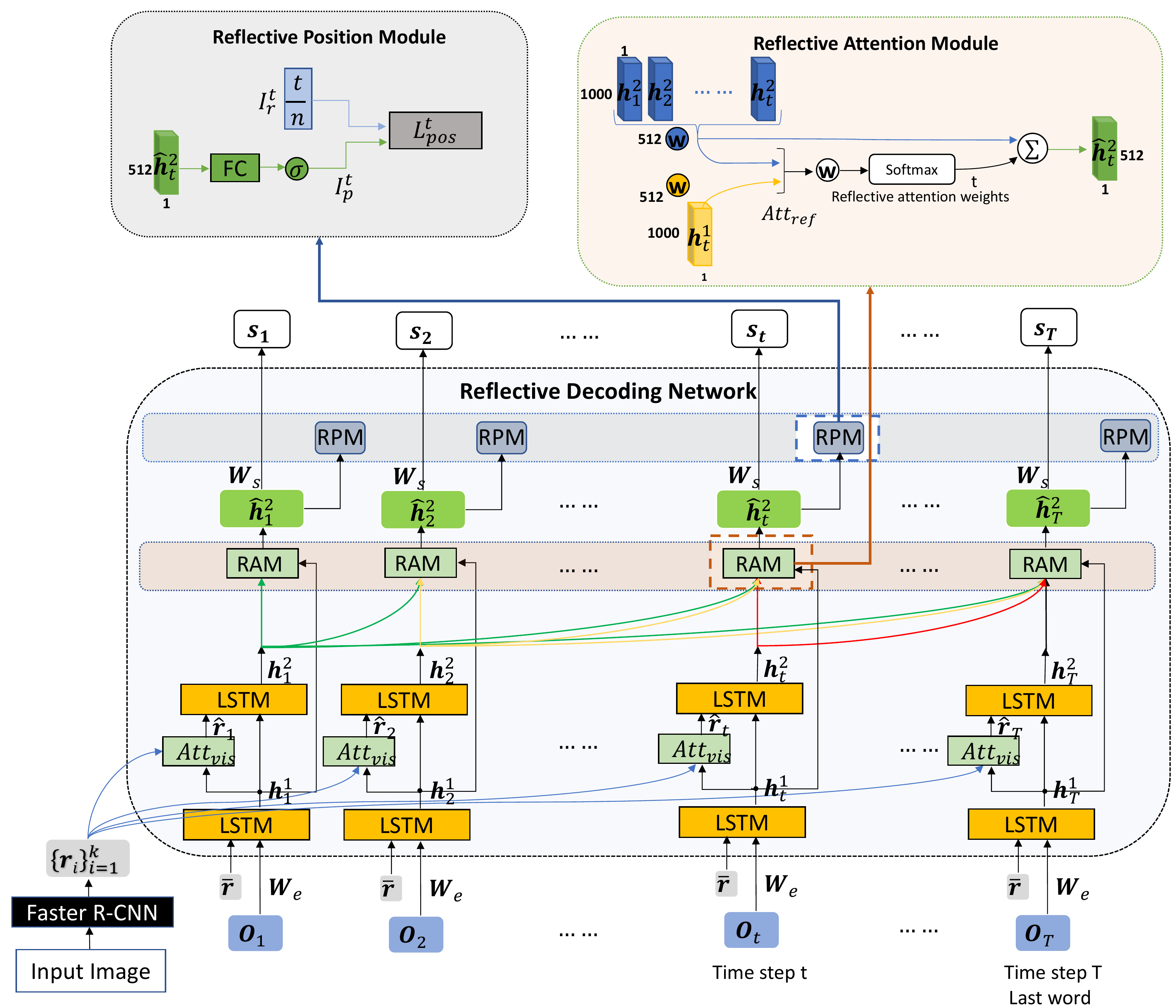}
	\caption{Overview of our framework. $Att_{ref}$ in RAM (Reflective Attention Module) is the attention layer used to selectively attend to the generated decoding hidden states, $Att_{vis}$ layer in Attention-based Recurrent Module determines the attention distribution over the detected image regions. $I_{p}^t$ and $I_{r}^t$ in RPM (Reflective Position Module) are respectively the $t$-th predicted and actual relative position in sentence.}
	\label{fig:model}
	\vspace{-0.3cm}
\end{figure*}

\paragraph{Image Captioning.}
State-of-the-art captioning methods are mostly driven by advancements in machine translation \cite{cho2014learning,sutskever2014sequence}, where the encoder-decoder framework has demonstrated to generate much more novel and coherent sentences compared to the traditional template-based~\cite{kulkarni2013babytalk,yang2011corpus} or search-based \cite{devlin2015language} methods.
In~\cite{donahue2015long,vinyals2015showtell}, the authors introduced a framework which utilizes a pre-trained CNN as an encoder to extract image features, followed by an RNN as a decoder to generate image descriptions.
This model was further improved by incorporating high-level semantic attribute information~\cite{wu2016value,yao2017boosting} or regularizing the RNN decoder \cite{chen2018regularizing}.
To distill the salient objects or important regions from an image, different kinds of attention mechanisms were integrated into the captioning framework to exam the relevant image regions when generating sentences~\cite{anderson2017bottom,lu2017knowing,xu2015showattendtell,yang2016review,you2016image}.

Fusion learning of multiple encoders or decoders forms an essential part of boosting image captioning performance.
In \cite{jiang2018recurrent}, the authors utilized multiple CNNs to extract complementary image features, which forms a more informative and integrated representation for decoder.
Yao \etal \cite{yao2018exploring} proposed GCN-LSTM to build two kinds of graphs to incorporate both semantic and spatial relations into the framework.
The outputs from two different separately trained decoders are linearly fused to produce the final prediction.

Similar to \cite{anderson2017bottom,lu2018neural,yao2018exploring}, our RDN also utilizes the attention mechanism and follows the encoder-decoder framework.
However, we explicitly study the coherence between words, which remedies the drawback of current captioning framework in modeling long-term dependency in decoder.

\smallskip\noindent\textbf{Language Attention in joint vision and language tasks.}
Learning language attention has attracted increasing attention in other joint vision and language problems, such as VQA and grounding referential expressions.
In \cite{lu2016hierarchical}, the authors proposed a model to jointly reason both visual and language attentions for visual question answering.
Yu \etal \cite{yu2018mattnet} attentively parsed the expressions into three phrase embeddings to address the task of referring expression comprehension.
Different from them, image captioning task is a sequential language generation process.
The target description of an image is unknown during inference stage.
So, our RDN explores the language attention based on the generated words in previous states. 
With more time steps, the attended language content will increase dynamically, which enables the word predicted later to capture more useful information for reference.

\vspace{-0.02in}
\smallskip\noindent\textbf{Language Attention in NLP tasks.}
Our RDN shares some ideas of the self-attention mechanism in machine translation models~\cite{miculicich2018self,tran2016recurrent,vaswani2017attention}, abstract summarization model~\cite{paulus2017deep} and dialogue system~\cite{mei2017coherent}. 
A typical self-attention model such as Transformer \cite{vaswani2017attention} aims to learn a latent representation for each position of a sequence by referring to the whole context.
In contrast, the Reflective Attention Module (RAM) of our RDN is designed as an attachable module which is seamlessly integrated into the recurrent decoding framework.
Thanks to our special two-layer recurrent structure, our RAM collaborates smoothly with the visual attention component of our RDN by sharing the same query value to optimize the captioning process jointly, which is beneficial to ensure our generated captions match with the visual content of an image.
To our knowledge, this paper is the first work in jointly exploring both visual and language attention in image captioning.
\vspace{-0.15in}


\section{Reflective Decoding Network}
\vspace{-0.05in}
The overall architecture of our framework is shown in Figure \ref{fig:model}.
Given an input image, our model first uses Faster R-CNN \cite{ren2015faster} as Encoder to obtain the visual features of objects in the image. 
The visual features are then fed to the our Reflective Decoding Network (RDN) to generate caption.
Our RDN contains three components:
(1) Attention-based Recurrent Module, which attends to the visual features from Encoder; 
(2) Reflective Attention Module, which employs textual attention to model the compatibility between current and past decoding hidden states, thus it is able to capture more historical and comprehensive information for word decision;
(3) Reflective Position Module, which introduces relative position information for each word in the generated caption and helps the model to perceive the syntactic structure of sentences.
RDN is able to tackle the long-term dependency difficulty in caption decoding. 

\vspace{-0.1in}

\subsection{Object-Level Encoder}
\vspace{-0.05in}
\label{sec:encoder} 
The encoder in encoder-decoder framework aims to extract meaningful semantic representation from an input image.
We leverage object detection module (Faster R-CNN~\cite{ren2015faster}) with pretrained ResNet-101~\cite{he2016deep} to produce the region-level representation.
The set of extracted regional visual representation $R_I$ of an image $I$ are denoted as $R_I$ = $\mathbf{\{r_i\}}^k_{i=1}$, $\mathbf{r}_i$$\in$$\mathbb{R}^D$, where $k$ denotes the number of extracted regions, $D$ denotes the feature dimension of each region, and $\mathbf{r}_i$ is the mean pooled convolutional feature within the extracted region.
Compared to the conventional uniform meshing method on CNN features, the object-level encoder focuses more on salient objects/regions in an image that is closely related to the perception mechanism in human visual system \cite{buschman2007top}.

\subsection{Reflective Decoder}
\label{sec:decoder} 
Given a set of regional image features $R_I$ produced by encoder, the goal for the decoder is to generate the caption $S$, where $S$ = $\{s_1,s_2,...,s_n\}$ consisting of $n$ words. 
The generated caption should not only capture the content information from the image but also be meaningful and coherent.
Specifically, in Figure~\ref{fig:model}, the Attention-based Recurrent Module is employed to selectively attend to the detected regional features and serves the basic function of a captioning decoder while Reflective Attention Module and Reflective Position Module are designed above it as assistants to further enhance captioning quality.
Thus, the complete Reflective Decoder is able to take both historical coherence between words and syntactic structure information into consideration while generating image captions.

Attention-based Recurrent Module includes the first LSTM layer and visual attention layer $Att_{vis}$, which is designed mainly for top-down visual attention calculation. 
Its input $\mathbf{x}_t^1$ at time step ${t}$ contains three concatenated parts, the mean-pooled image feature  $\bar{\mathbf{r}}$ = $\frac{1}{k}\sum_{i=1}^{k}\mathbf{r}_i$, the embedding vector ${\mathbf{W}_e}\mathbf{O}_{t}$ for current input word $\mathbf{O}_{t}$ and the previous output $\mathbf{h}_{t-1}^{2}$ from the second LSTM layer, where $\bar{\mathbf{r}}$ represents the contextual information of the given image, ${\mathbf{W}_e}$$\in$$\mathbb{R}^{E \times D_o}$ is the embedding matrix for the one-hot vector $\mathbf{O}_{t}$, $D_o$ is the size of the captioning vocabulary and $E$ is the embedding size.
The formula for updating the LSTM units in the first layer is defined as :
\begin{equation}
\mathbf{h}_t^1 = {\rm LSTM}(\mathbf{x}_t^1,\mathbf{h}_{t-1}^1),\quad
\mathbf{x}_t^1 = [\bar{\mathbf{r}}, {\mathbf{W}_e}\mathbf{O}_{t},\mathbf{h}_{t-1}^{2}].
\end{equation}
For the visual attention layer $Att_{vis}$, given the generated $\mathbf{h}_t^1$ and the set of $k$ image features $R_I$ = $\{\mathbf{r}_i\}^k_{i=1}$, we calculate the normalized attention weight $\alpha_{t}^{vis}$ distribution over all the proposed object-level region denotes as :
\begin{equation}
\alpha_{i,t}^{vis} = \mathbf{W}_v^1{\rm tanh}(\mathbf{W}_{rv}^1\mathbf{r}_i+\mathbf{W}_{hv}^1{\mathbf{h}_t^1}),
\end{equation}
\begin{equation}
\alpha_t^{vis} = {\rm softmax}(a_t^{vis}),\quad
a_t^{vis} =\left \{\alpha_{i,t}^{vis} \right \}_{i=1}^k,
\end{equation}
where $\mathbf{W}_v^1$ $\in$ $\mathbb{R}^{1 \times D_{v}}$, $\mathbf{W}_{rv}^1$ $\in$ $\mathbb{R}^{D_v \times D_R}$, $\mathbf{W}_{hv}^1$$\in$ $\mathbb{R}^{D_v \times D_{h}}$ are learned embedding matrices, $\alpha_t^{vis}$ denotes the calculated attention probability for each regional feature $\mathbf{r}_i$ at time step $t$.
So the attended feature is the weighted combination of each subregion, $\hat{\mathbf{r}}_t$ = $\sum_{i=1}^{k}\alpha_{i,t}^{vis}\mathbf{r}_i$ based on the weight distribution parameter $a_t^{vis}$.
\vspace{-3mm}
\subsubsection{Reflective Attention Module.}
\label{sec:attention_module} 
The Reflective Attention Module contains reflective attention layer $Att_{ref}$, combined with the second layer of LSTM, which is designed to output language description. 
Its input vector is concatenated by the attended feature result $\hat{\mathbf{r}}_t$ and the hidden state $\mathbf{h}_t^1$ .
Thus the formula for updating the LSTM units in the second layer of LSTM is denoted as :
\begin{equation}
\mathbf{h}_t^2 = {\rm LSTM}(\mathbf{x}_t^2,\mathbf{h}_{t-1}^2),\quad
\mathbf{x}_t^2 = [\hat{\mathbf{r}}_t, \mathbf{h}_t^1].
\end{equation}
Based on the current hidden state $\mathbf{h}_t^2$ at the time step $t$ and the past hidden states set $\{\mathbf{h}_1^2,\mathbf{h}_2^2,...,\mathbf{h}_{t-1}^2\}$, the reflective attention layer $Att_{ref}$ calculates the normalized weight distribution $\alpha_{t}^{ref}$ above all the generated $t$ hidden states as shown in the top right of Figure~\ref{fig:model}. 
The formula is defined as : 
\begin{equation}
\alpha_{i,t}^{ref} = \mathbf{W}_h^2{\rm tanh}(\mathbf{W}_{h_2h}^2\mathbf{h}_i^2+\mathbf{W}_{\mathbf{h}_1h}^2{\mathbf{h}_t^1}),
\end{equation}
\begin{equation}
\alpha_t^{ref} = {\rm softmax}(a_t^{ref}),\quad
a_t^{ref} =\left \{\alpha_{i,t}^{ref} \right \}_{i=1}^t,
\end{equation}
where $\mathbf{W}_h^2$ $\in$ $\mathbb{R}^{1 \times D_{f}}$, $\mathbf{W}_{h_2h}^2$ $\in$ $\mathbb{R}^{D_f \times D_h}$, $\mathbf{W}_{h_1h}^2$ $\in$ $\mathbb{R}^{D_f \times D_{h}}$ are three trainable matrices parameters, $\alpha_t^{ref}$ denotes the generated attention probability set for each hidden state $\mathbf{h}_i$ in the set  $\{\mathbf{h}_i^2\}^t_{i=1}$ at time step $t$ and $\alpha_{i,t}^{ref}$ reflects the relevance between the past predicted word at $i$-th step and current prediction (at $t$-th step) by measuring the compatibility between their corresponding hidden states.
So we can calculate the attended hidden state result $\hat{\mathbf{h}}_t^2$ = $\sum_{i=1}^{t}\alpha_{i,t}^{ref}\mathbf{h}_i^2$.

The reflective decoding output $\hat{\mathbf{h}}_t^2$ of the top attention layer $Att_{ref}$ is utilized to predict the word $s_{t}$ under the conditional probability distribution :
\begin{equation}
\label{con:prob}
p(s_t|s_{1:t-1}) = {\rm softmax}(\mathbf{W}_s\hat{\mathbf{h}}_t^2 + \mathbf{b}_s),
\end{equation}
where $\mathbf{W}_s$ $\in$ $\mathbb{R}^{D_{o} \times D_h}$ are the trainable weights and $\mathbf{b}_s$ $\in$ $\mathbb{R}^{D_{o}}$ are the biases.
By calculating $s_{t}$ in this way, all the generated hidden states $\{\mathbf{h}_i\}_{i=1}^t$ play a role in word precision and their extent of contributions can be clearly visualized, as will be demonstrated in section~\ref{sec:qualitative_analysis}.

It should be noted that our proposed Reflective Attention Module models the dependencies between pairs of words at different time steps explicitly, taking into account the corresponding hidden states.
In contrast, LSTM memorizes the historical sequence information by balancing the overall relevance of all time steps instead of modeling the dependency for each pair of words specifically. 

\vspace{-0.15in}
\subsubsection{ Reflective Position Module.}
\vspace{-0.08in}
It is often the case that many of the words have relatively fixed positions in a sentence due to the syntactic structure in natural language.
For example, the numeral and subject words,~\ie `a man' or `a woman', mostly appear at the beginning of the sentence while the predicates tend to occupy the middle position.
So we propose the Reflective Position Module by injecting the word position information during training as a guidance for the sequence decoding model to perceive its relative position or progress in the whole sentence.
When decoding the $t$-th word, its actual relative position $I^t_{r}$ and the predicted relative position $I^t_{p}$ are calculated as :
\vspace{-0.02in}
\begin{equation}
\vspace{-0.02in}
I^t_{r} = \frac{t}{n},\quad I^t_{p} = \sigma(\mathbf{W}_l\hat{\mathbf{h}}_t^2),
\label{eq:eq8}
\end{equation}
where $n$ is the length of the sentence, $\sigma$ is the sigmoid function and $\mathbf{W}_l$$\in$ $\mathbb{R}^{1 \times D_{h}}$ is the trainable relative position embedding matrix, respectively.
The reflective position module shown in top left of Figure~\ref{fig:model} aims to minimize the difference between $I^t_{r}$ and $I^t_{p}$ by designing a loss function, which refines the attended hidden state $\hat{\mathbf{h}}_t^2$ mentioned in~\ref{sec:attention_module} by enabling it to perceive more sequential information of its relative position.

It is different from the popular position embedding methods~\cite{gehring2017convolutional, vaswani2017attention}, which add the absolute position embedding to the corresponding input features in each dimension. Our Reflective Position Module models the relative position information individually in a supervised way.
A key benefit of this design is that it can avoid the potential inter-pollution between the regular input feature and the position embedding, and equip our model with a strong perception of relative position for each word in the caption. Thus, the syntactic structure in natural language can be well preserved.

\subsection{Training and Inference}
\label{sec:training_inference}

\paragraph{Training.} 
Two kinds of losses are utilized for optimizing our RDN model.
The first is the cross entropy loss in traditional captioning training, which is to minimize the negative log likelihood:
\vspace{-0.05in}
\begin{equation}
L_{\rm xe} = -{\rm log}p(S^*|I) = - \sum_{t=2}^{n}{\rm log}p(s_t^*|s_{1:t-1}^*),
\end{equation}
where $I$ is the given image, $S^*$ is the ground truth caption, formula for calculating $p(s_t^*|s_{1:t-1}^*)$ is defined in equation \ref{con:prob} and $s_0^*$ is the start of the sentence. 

The second loss is defined as the Position-Perceptive Loss $L_{pos}$:
\vspace{-0.05in}
\begin{equation}
L_{pos} = \sum_{t=1}^{n} \left \| I^t_{r}-I^t_{p} \right \|^2, 
\end{equation}
where $I^t_{r}$ and $I^t_{p}$ are the actual relative position and predicted relative position defined in Equation \ref{eq:eq8} and $L_{pos}$ is designed to minimize the gap between them. 

The objective function for optimizing our RDN is defined as :
\begin{equation}
L = L_{\rm xe}+ \lambda L_{pos}.
\end{equation}
The trade-off parameter $\lambda$ balances the contribution between the traditional caption loss in encoder-decoder framework and the Position-Perceptive Loss.
\vspace{-3mm}
\paragraph{Inference.} 
During the inference stage, since the length of the whole predicted sentence is unknown, the relative position information is removed from the input.
As the discrepancy problem~\cite{lamb2016professor} between training and inference, which means the previous ground truth captioning token is not available for inference, we use the previously predicted word as input instead of ground truth word as in~\cite{chen2015mind,xu2015showattendtell}.
This method is called teacher forcing algorithm~\cite{williams1989learning}.
Also, we adopt the popular beam search strategy which iteratively selects the top-k best sentences at time step $t$ as candidates to generate the new top-k sentence at time $t+1$ in our experiment instead of greedy search.
\vspace{-0.02in}

\begin{table*}[t]
	\centering
	\tabcolsep=0.15cm
	\renewcommand{\arraystretch}{0.9}
	{\small
		\begin{tabular}{ l | c c c c c c c c }
			\toprule
			Model & BLEU-1 & BLEU-2  & BLEU-3  & BLEU-4  & METEOR & ROUGE-L & CIDEr & SPICE \\
			\midrule
			Baseline    & 77.0 & 61.3 & 47.2 & 36.1  & 26.8 & 56.1 & 113.2 & 20.1 \\
			RDN$_{pos}$   & 77.4 & 61.6 & 47.5 & 36.3  & 27.0 & 56.5 & 114.3 & 20.4 \\
			RDN$_{ref} $   & \textbf{77.6} & 61.6 & 47.4 & 36.3  & 27.1 & 56.7 & 115.0 & \textbf{20.5} \\
			RDN   & 77.5 & \textbf{61.8} & \textbf{47.9} & \textbf{36.8} & \textbf{27.2} & \textbf{56.8} & \textbf{115.3} & \textbf{20.5} \\
			\bottomrule
		\end{tabular}
	}
    \vspace{0.1cm}
	\caption {Ablation study on COCO `Karpathy' test split on single model. Our experiments show the contribution for reflective attention and position module, respectively. Results are obtained with beam size 5 without CIDEr optimization. All value reported in percentage (\%).
	}
    \vspace{-0.1in}
	\label{table:ablation_study}
\end{table*}

\begin{table*}[t]
	\centering
	\tabcolsep=0.15cm
	\renewcommand{\arraystretch}{0.9}
	{\small
		\begin{tabular}{ l | c c c c c c c c }
			\toprule
			Model & BLEU-1 & BLEU-2  & BLEU-3  & BLEU-4  & METEOR & ROUGE-L & CIDEr & SPICE \\
			\midrule
			Review Net~\cite{yang2016review}  & - & - & - & 29.0  & 23.7 & - & 88.6 & - \\
			LSTM-A3~\cite{yao2017boosting}  & 73.5 & 56.6 & 42.9 & 32.4  & 25.5 & 53.9 & 99.8 & 18.5 \\
			Att2in \cite{rennie2017self}  & - & - & - & 31.3  & 26.0 & 54.3 & 101.3 & - \\
			Adaptive~\cite{lu2017knowing}  & 74.2 & 58.0 & 43.9 & 33.2  & 26.6 & - & 108.5 & - \\
			Up-Down~\cite{anderson2017bottom}  & 77.2 & - & - & 36.2  & 27.0 & 56.4 & 113.5 & 20.3 \\
			RFNet~\cite{jiang2018recurrent}  & 76.4 & 60.4 & 46.6 & 35.8  & \textbf{27.4} & 56.5 & 112.5 & \textbf{20.5} \\
			\midrule
			RDN   & \textbf{77.5} & \textbf{61.8} & \textbf{47.9} & \textbf{36.8} & 27.2 & \textbf{56.8} & \textbf{115.3} & \textbf{20.5} \\
			\bottomrule
		\end{tabular}
	}
	\vspace{0.1cm}
	\caption {Performance comparison on MSCOCO `Karpathy' test split on single model. All image captioning models trained without optimizing CIDEr metric. $(-)$ indicates the metric is not provided.  
	}
	\vspace{-0.1in}
	\label{table:single_karpathy}
\end{table*}

\begin{table*}[!t]
	\centering
	\tabcolsep=0.15cm
	\renewcommand{\arraystretch}{0.9}
	{\small
		\begin{tabular}{ l | c c c c c c c c }
			\toprule
			Model & BLEU-1 & BLEU-2  & BLEU-3  & BLEU-4  & METEOR & ROUGE-L & CIDEr & SPICE \\
			\midrule
			NIC~\cite{vinyals2015showtell} & - & - & - & 32.1  & 25.7 & - & 99.8 & - \\
			Att2in~\cite{rennie2017self}  & - & - & - & 32.8  & 26.7 & 55.1 & 106.5 & - \\
			Review Net~\cite{yang2016review}  & 76.7 & 60.9 & 47.3 & 36.6  & 27.4 & 56.8 & 113.4 & 20.3 \\
			RFNet~\cite{jiang2018recurrent}  & 77.4 & 61.6 & 47.9 & 37.0  & \textbf{27.9} & 57.3 & 116.3 & \textbf{20.8} \\
			\midrule
			RDN & \textbf{77.6} & \textbf{62.2} & \textbf{48.6} & \textbf{37.8}  & 27.5 & \textbf{57.4} & \textbf{117.3} & 20.6 \\
			\bottomrule
		\end{tabular}
	}
	\vspace{0.1cm}
	\caption {Performance comparison on MSCOCO `Karpathy' test split on ensemble models trained with cross entropy loss. Our model is the ensembling result of 6 single models initialized with different random seeds.
	}
	\vspace{-0.1in}
	\label{table:ensemble}
\end{table*}


\section{Experiments}
\vspace{1mm}
\subsection{Datasets and Experimental Settings}

\smallskip\noindent\textbf{COCO Dataset.}
COCO captions dataset \cite{chen2015microsoft}  contains 82,783 images for training and 40,504 images for validation.
Each image has five corresponding human-annotated captions.
Also, we adopt the `Karpathy' splits setting \cite{karpathy2015deep}, which includes 113,287 training images, 5K validation images and 5K testing images for offline evaluation.
For the online server evaluation, the entire images and captions in dataset is used for training.
Following the text preprocessing in \cite{anderson2017bottom}, we convert all the captions to lower case and remove the less frequent words which occur less than 5 times, obtaining a captioning vocabulary of 10,010 words.


\smallskip\noindent\textbf{Visual Genome Dataset.}
Visual Genome \cite{krishna2017visual} is a large dataset for modeling the interactions and relations between objects within an image.
The dataset consists of 108K images with densely annotated objects, attributes and pairwise relations.
Compared to \cite{yao2018exploring}, we only utilize the annotated object and attribute data from the dataset to pretrain the object-level encoder and discard the pairwise relation data. 
We follow the same data split setting in \cite{anderson2017bottom} to include 98K images for training, 5K images respectively for validation and testing.
After cleaning these annotated object and attribute strings, we obtain a dataset including 400 attributes and 1,600 objects classes to train our Faster R-CNN model.

\smallskip\noindent\textbf{Evaluation Metrics.}
To objectively evaluate the performance of our captioning model, we use five widely accepted automatic evaluation metrics, including CIDEr~\cite{vedantam2015cider}, SPICE~\cite{anderson2016spice}, BLEU~\cite{papineni2002bleu}, METEOR~\cite{denkowski2014meteor} and ROUGE-L~\cite{lin2004rouge}.


\smallskip\noindent\textbf{Implementation Details.}
We implement our RDN using Caffe~\cite{jia2014caffe}.
To train the object-level encoder, we use the Faster R-CNN with ResNet-101 pre-trained for image classification on ImageNet~\cite{russakovsky2015imagenet} and further refine it on the Visual Genome dataset.
For each image, we set the IoU thresholds for region proposal suppression and object prediction to 0.7 and 0.3 respectively.
For the remaining image subregions, we set a filter threshold 0.2. 
We rank the leftover boxes by their confidence scores from high to low and choose no more than top 100 as the final feature representations.
Each region with dimension number 2,048 is the global average pooling result of the layer Res5c.

We set the word embedding size and the hidden size in each LSTM layer to 1,000.
The dimensions for attention layers $Att_{vis} $ and $Att_{ref}$ are set to 512 respectively.
During training, the initial learning rate is set to 0.01 and the polynomial decay strategy is adopted to decline the effective learning rate to zero by 70k iterations using a batch size 100.
We tune the trade-off parameter $\lambda$ on the `Karpathy' validation split to obtain the best performance and finally set it to 0.02.
For data augmentation, we adopt it only during the online test server submission to boost performance by flipping the original image and randomly cropping 90\%. 
During decoding process, the beam search size is set to 5.

\begin{table*}[t]
	\centering
	\tabcolsep=0.15cm
	\renewcommand{\arraystretch}{0.9}
	{\small
		\begin{tabular}{ l | c | c | c | c | c | c | c | c | c | c | c | c }
			\toprule
			\multirow{2}{*}{Model} &
			\multicolumn{2}{c|}{BLEU-1} &
			\multicolumn{2}{c|}{BLEU-4} &
			\multicolumn{2}{c|}{METEOR} &
			\multicolumn{2}{c|}{ROUGE-L} &
			\multicolumn{2}{c|}{CIDEr} &
			\multicolumn{2}{c}{SPICE} \\
			\cline{2-13}
			& c5 & c40 & c5 & c40 & c5 & c40 & c5 & c40 & c5 & c40 & c5 & c40 \\
			\hline
			NIC \cite{vinyals2015showtell}  & 71.3 & 89.5 & 30.9 & 58.7 & 25.4 & 34.6 & 53.0 & 68.2  & 94.3 & 94.6 & 18.2 & 63.6 \\
			Review Net~\cite{yang2016review}  & 72.0 & 90.0 & 31.3 & 59.7 & 25.6 & 34.7 & 53.3 & 68.6  & 96.5 & 96.9 & 18.5 & 64.9 \\
			LSTM-A3~\cite{yao2017boosting}  & 78.7 & 93.7 & 35.6 & 65.2 & 27.0 & 35.4 & 56.4 & 70.5  & 116.0 & 118.0 & - & - \\
			Adaptive~\cite{lu2017knowing}  & 74.8 & 92.0 & 33.6 & 63.7 & 26.4 & 35.9 & 55.0 & 70.5  & 104.2 & 105.9 & 19.7 & 67.3 \\
			Att2all~\cite{rennie2017self}  & 78.1 & 93.7 & 35.2 & 64.5 & 27.0 & 35.5 & 56.3 & 70.7  & 114.7 & 116.7 & - & - \\
			Up-Down~\cite{anderson2017bottom}  & 80.2$_{\textcolor{red}{2}}$ & 95.2$_{\textcolor{red}{2}}$ & 36.9$_{\textcolor{red}{3}}$ & 68.5$_{\textcolor{red}{3}}$ & 27.6$_{\textcolor{red}{3}}$ & 36.7$_{\textcolor{red}{3}}$ & 57.1$_{\textcolor{red}{3}}$ & 72.4$_{\textcolor{red}{3}}$  & 117.9$_{\textcolor{red}{3}}$ & 120.5$_{\textcolor{red}{3}}$ & - & - \\
			RFNet~\cite{jiang2018recurrent}  & 80.4$_{\textcolor{red}{1}}$ & 95.0$_{\textcolor{red}{3}}$ & 38.0$_{\textcolor{red}{1}}$ & 69.2$_{\textcolor{red}{2}}$ & 28.2$_{\textcolor{red}{1}}$ & 37.2$_{\textcolor{red}{2}}$ & 58.2$_{\textcolor{red}{1}}$ & 73.1$_{\textcolor{red}{2}}$  & 122.9$_{\textcolor{red}{1}}$ & 125.1$_{\textcolor{red}{2}}$ & - & - \\
			\midrule
			RDN & 80.2$_{\textcolor{red}{2}}$ & 95.3$_{\textcolor{red}{1}}$ & 37.3$_{\textcolor{red}{2}}$ & 69.5$_{\textcolor{red}{1}}$ & 28.1$_{\textcolor{red}{2}}$ & 37.8$_{\textcolor{red}{1}}$ & 57.4$_{\textcolor{red}{2}}$ & 73.3$_{\textcolor{red}{1}}$  & 121.2$_{\textcolor{red}{2}}$ & 125.2$_{\textcolor{red}{1}}$ & - & - \\
			\bottomrule
		\end{tabular}
	}
	\vspace{0.1cm}
	\caption{Performance comparison of published image captioning models on COCO Leaderboard. RDN achieves superior performance when comparing to other state-of-the-art methods. Top-3 rankings are indicated by red footnote for each metric.
	}
	\label{table:leaderboard}
	\vspace{-0.2in}
\end{table*}

\subsection{Ablation Study on Reflective Modules}

To study the effects of Reflective Attention Module and Reflective Position Module in our model, an ablation experiment is designed to compare the performance with following combinations:
(1) Baseline: the baseline denotes the RDN without Reflective Attention Module and Reflective Position Module;
(2) RDN$_{pos}$: the RDN with the Reflective Attention Module removed, with only position module reserved, the number of attention layers in decoder is reduced to one;
(3) RDN$_{ref}$: the RDN with the Reflective Position Module removed, cutting down the relative position information input;
(4) RDN: the complete RDN implementation.

In Table~\ref{table:ablation_study}, it can be observed that both the Reflective Position Module and Reflective Attention Module are important for our model and RDN improves the caption performance over all the metrics compared to baseline.
The fact that RDN$_{pos}$ outperforms baseline model validates the contribution of Reflective Position Module to enhance the quality of decoding hidden states during caption generation.
Also, by injecting the Reflective Attention Module, RDN$_{ref}$ performs obviously better than baseline, which shows the importance of model's ability to capture long-term dependency between words.
In particular, with a suitable combination of the two modules, RDN achieves the best result with CIDEr score 115.7, BLEU-4 score 37.0 and BLEU-3 score 47.9, yielding the improvement over our baseline model by 2.0\%, 2.2\% and 1.5\% respectively, which is a considerable advancement over the benchmark. 
Compared the baseline model with total 1.15B parameters, RDN has only 0.84\% more in model size, which is neglectable.

\subsection{Performance Comparison and Analysis}
\label{sec:performance}

We compare our proposed RDN with other state-of-the-art image captioning methods considering different aspects both in offline and online situation.
Latest and representative works include:
(1) \textbf{Adaptive}~\cite{lu2017knowing} which proposes the adaptive attention through designing a visual sentinel gate for captioning model to decide whether to attend to the image feature or just rely on the sequential language model,
(2) \textbf{LSTM-A3}~\cite{yao2017boosting} which incorporates the high level semantic attribute information to the encoder-decoder model,
(3) \textbf{Up-Down}~\cite{anderson2017bottom} which introduces the bottom-up and top-down attention mechanism  to enable attention calculated at the level of objects or salient subregions
and (4) \textbf{RFNet}~\cite{jiang2018recurrent} which uses multiple kinds of CNNs to extract complementary image feature and generate a more informative representation for the decoder.

For fair comparison, our model and the baseline use standard ResNet-101 as basic architecture for encoder and all the reported results on test portions of MSCOCO `Karpathy' splits are trained without additional CIDEr optimization~\cite{rennie2017self}.
GCN-LSTM \cite{yao2018exploring} is not included because it uses the additional densely annotated pair-wise relation data between objects to pretrain semantic relation detector and build convolutional graphs.
We only adopt CIDEr optimization strategy for the online server submission, since directly optimizing the CIDEr metric has little effect on perceived caption quality during human evaluation~\cite{liu2016optimization} and small difference in its optimization implementation would influence the caption performance a lot.

\begin{figure}[tb]
	\centering
	\includegraphics[width=1.0\linewidth]{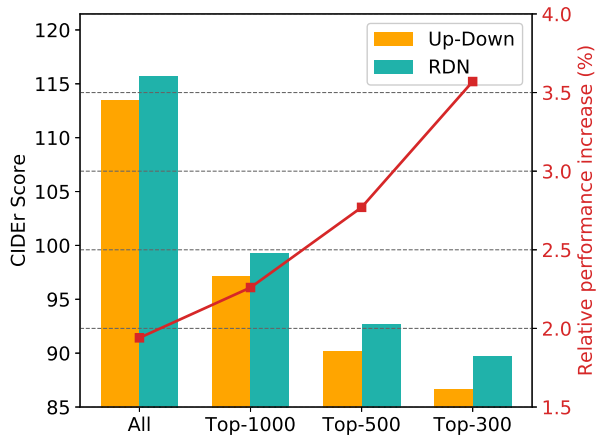}
	\caption{Performance comparison between our RDN model and Up-Down~\cite{anderson2017bottom} on hard Image Captioning as a function of average length of annotations (ground truth captions). We rank the `Karpathy' test set according to their average length of annotations in descending order and extract four different size of subsets. Smaller subset corresponds to averagely longer annotations and harder captioning. It reveals that our model exhibits more superiority over Up-Down in harder cases. 
	}
	\label{fig:bar}
	\vspace{-0.25in}
\end{figure}

\vspace{-0.1in}
\subsubsection{Quantitative Analysis}
\label{sec:quantitative_analysis}
\paragraph{Offline Evaluation.}
For offline evaluation, we compare performance of different models on `Karparthy' split dataset both in single and ensemble model situations. 
In Table \ref{table:single_karpathy}, it can be observed that our single RDN achieves the best results among all existing captioning methods across the six evaluation metrics including all the BLEU entries, ROUGE-L and CIDEr, performs on par with RFNet in SPICE and is slightly inferior to it in METEOR.
Different from the previous  captioning models (Up-Down, RFNet, Review Net, Adaptive, etc.) that boost performance through extracting more indicative and compact visual representation, the enhancement of our captioning model only attributes to a better reasoning and inference ability of the decoder directly on the target side.
Moreover, from the Table~\ref{table:ensemble}, we can see that our ensembled RDN outperforms other ensemble models in most of evaluation metrics, with the highest CIDEr score 117.3, and performs only inferiorly to RFNet in METEOR and SPICE entry.
RDN is the ensembling result of 6 single models with different random seed initialization while RFNet is composed of 4 RFNets with a total of 20 groups of different image representations.


\begin{figure*}[!t]
	\centering
	\includegraphics[width=0.9\linewidth]{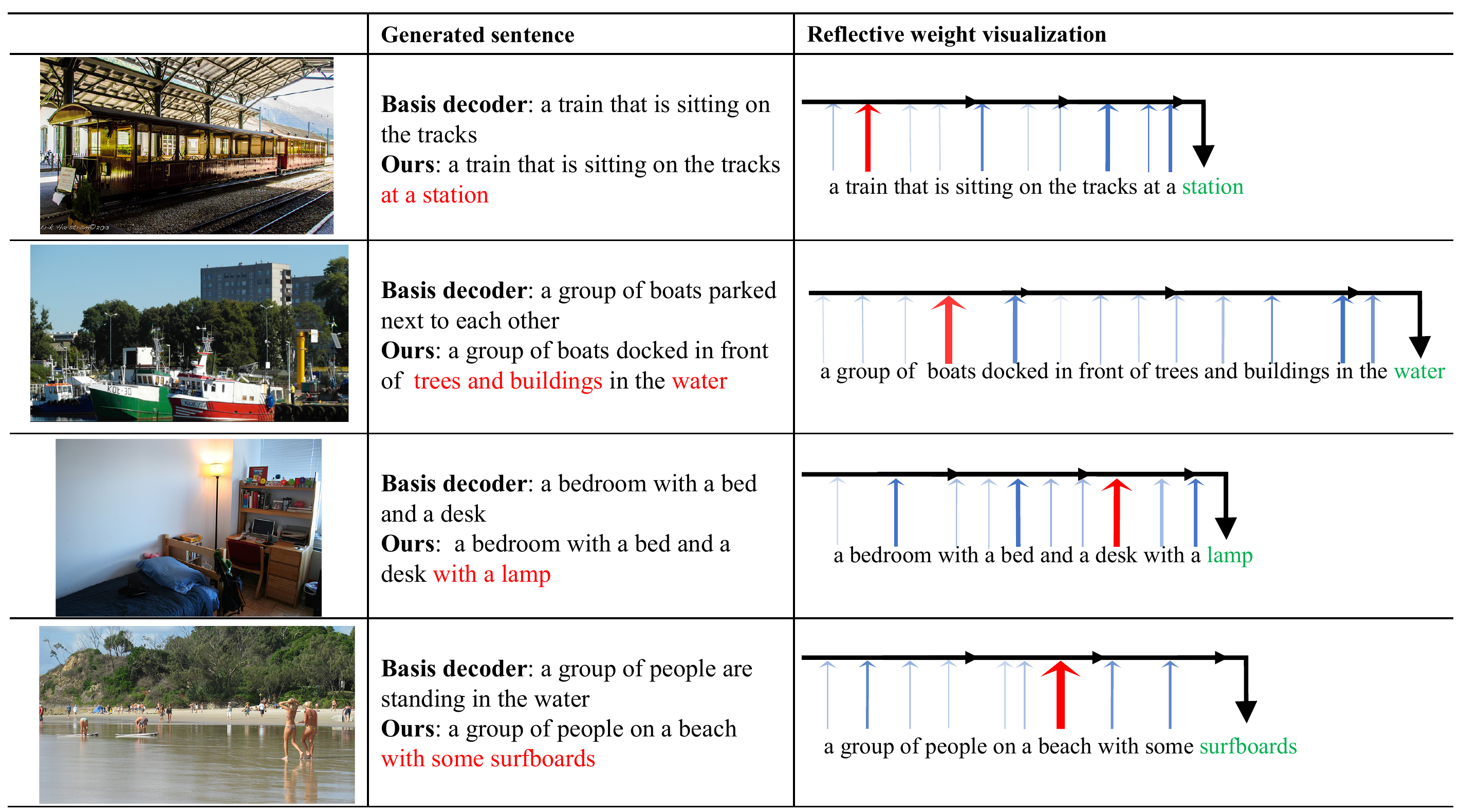}
	\caption{Examples of captions generated by our RDN compared to the basis decoder (using traditional LSTM) and their reflective attention weight distribution over the past generated hidden states when predicting the key words highlighted in green. The thicker line indicates a relatively larger weight and the red line means the largest contribution. More examples are provided in the supplementary material.}
	\label{fig:example2}
	\vspace{-0.2in}
\end{figure*}

\vspace{-0.04in}
\smallskip\noindent\textbf{Online Evaluation on COCO Testing Server.} 
We also compare our model with the published state-of-the-art captioning systems on COCO Testing Server with 5(c5) and 40(c40) reference sentences as shown in Table~\ref{table:leaderboard}.
Using the ensemble of 9 CIDEr optimized models, our RDN achieves leading performance over all metrics while performing on par with RFNet~\cite{jiang2018recurrent}.
Surprisingly, RFNet has a much better performance in online evaluation compared to the offline case, in which it performs much poorer than our model (even poorer than Up-Down~\cite{anderson2017bottom} in some cases) shown in Table~\ref{table:single_karpathy}. 
Since the code of RFNet is not released, it is hard to investigate the inconsistence. 
Nevertheless, our RDN achieves the superior performance in all the c40 entries.
Compared to c5, c40 has far more reference sentences and existing evaluation experiments show it achieves higher correlation with human judgement~\cite{chen2015microsoft,vedantam2015cider}. 
Moreover, our model is more simple and elegant with only one encoder-decoder in single model compared to RFNet, which utilizes multiple encoders (ResNet, DenseNet \cite{huang2017densely}, Inception-V3 \cite{szegedy2016rethinking} , Inception-V4, and Inception-ResNet-V2 \cite{szegedy2017inception})  to extract 5 groups of features and includes time-consuming feature fusion steps to produce the final thought vectors.
Besides, our RDN boosts captioning performance by optimizing the decoding stage while RFNet mainly focuses on improving the encoder.
Thus, it is a promising extension to apply the encoding mechanisms of RFNet to our RDN.
Compared to Up-Down~\cite{anderson2017bottom}, which uses traditional LSTM and object-level encoder, the CIDER-c40, CIDER-c5 and METEOR-c40 are improved by 3.9\%, 2.7\% and 2.9\%.

\vspace{-0.04in}
\smallskip\noindent\textbf{Evaluation on hard Image Captioning.}
We further investigate the effect of the average length of annotations (ground truth captions) on the captioning performance, since generally the images with averagely longer annotations contain more complex scenes and thus are harder for captioning.
Specifically, we rank the whole `Karparthy' testset (5000 images) according to their average length of annotations in descending order and extract four different size of subsets (all set, top-1000, top-500, top-300 respectively). Smaller subset corresponds to averagely longer annotations and implies harder image captioning. Figure~\ref{fig:bar} shows the comparison between our RDN and Up-Down~\cite{anderson2017bottom} (main difference of the two models is that Up-Down uses traditional LSTM). 
It reveals that the performance of both models are decreasing with the increasing average length of annotations, which reflects that the captioning is getting harder. However, our model exhibits more superiority over Up-Down in harder cases, which in turn validates the ability of our RDN to capture the long-term dependencies within captions.
We also provide one such comparison between our RDN and Att2in~\cite{rennie2017self} on hard captioning in the supplementary file.


\vspace{-0.15in}
\subsubsection{Qualitative Analysis}
\label{sec:qualitative_analysis}
To investigate the physical interpretability of RDN model's improvement, some qualitative comparisons of captioning results are shown in Figure~\ref{fig:example2}. 
Compared to basis decoder, our RDN is able to generate more detailed and discriminative descriptions for images.
Take the first case in Figure~\ref{fig:example2} as an example, the basis decoder can provide a general and reasonable caption for the image.
However, it cannot recognize the word `station' which actually exists in the image while our model successfully infers it based on the previously generated words, especially the closely related words `train',`tracks' and `sitting'.
For the reflective weight visualization, the generated words with the largest contribution to the predicting word are usually strong related in vocabulary coherence, such as the correlations between words ``boat'' and ``water'', ``beach'' and ``surfboards''.
We show additional results in supplementary material.
\vspace{0.02in}

\begin{figure}[t]
	\centering
	\includegraphics[width=0.92\linewidth]{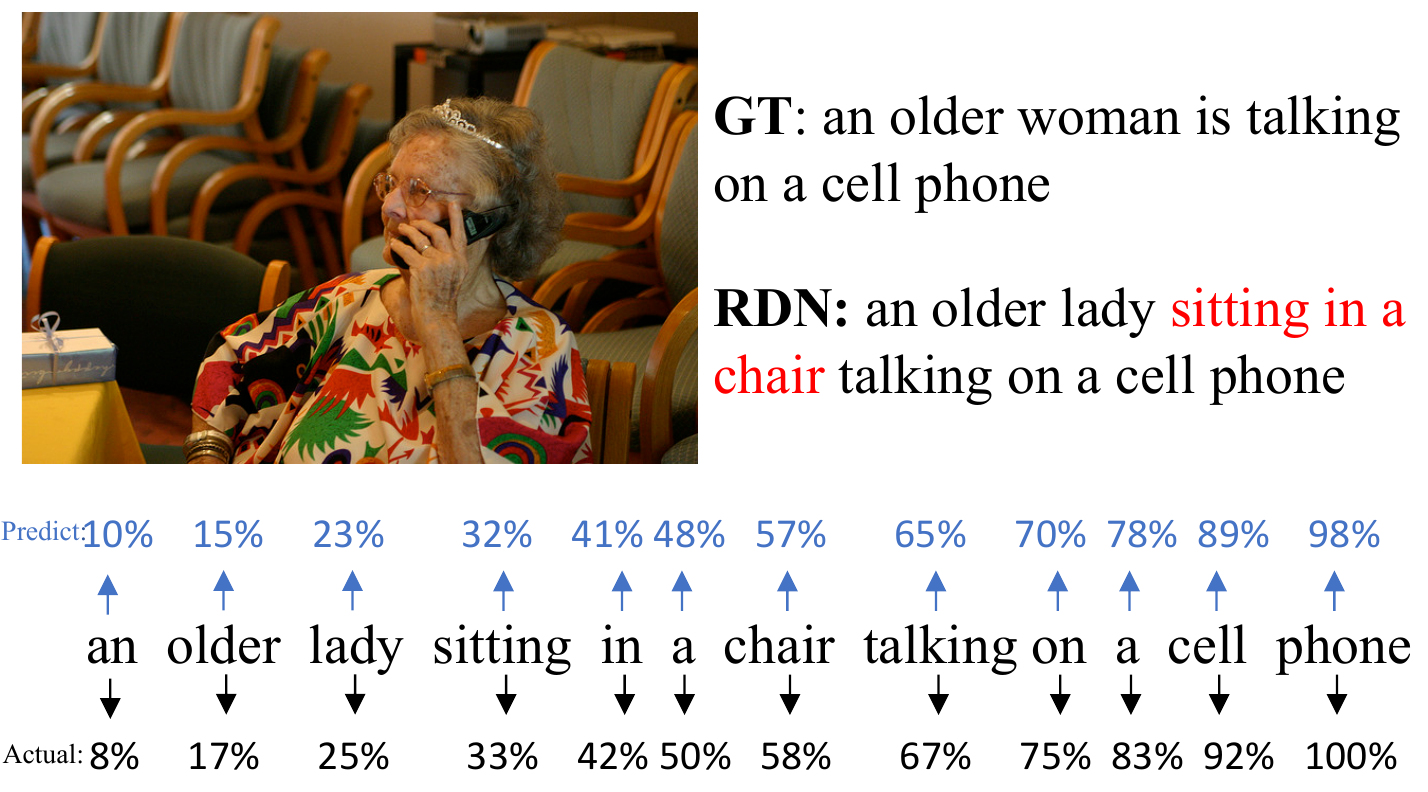}
	\caption{\textbf{Top}: Ground-truth caption and the example caption generated by our RDN model. \textbf{Bottom}: The predicted relative position value (shown in blue) from the Reflective Position Module and the actual relative position for each word in the sentence. All value reported in integer percentage value (\%). }
	\label{fig:example3}
	\vspace{-0.1in}
\end{figure}

In Figure~\ref{fig:example3}, compared to other captioning models, our RDN is able to predict the word and its relative position in the sentence simultaneously during caption generation.
The predicted relative position in blue for each word is highly close to its actual relative position value in sentence, which demonstrates a good position-perceptive ability of our model to capture the syntactic structure of a sentence.

\vspace{-2mm}

\section{Conclusion}
We have presented a novel architecture, Reflective Decoding Network (RDN), which explicitly explores the coherence between words in the captioning sentence and enhances the long-term sequence inference ability of LSTM.
Particularly, the attention mechanism applied in both visual and textual domain and the proposed position-perceptive scheme are to maximize the reference information available for captioning model.
We also show how the learned attention in textual domain can provide interpretability during the captioning generation process from a new perspective.
Extensive experiments conducted on standard and hard COCO image captioning dataset with superior performance validate the effectiveness of our proposal.
For future work, we are interested in extending our model to source code captioning and text summarization.

{\small
	\bibliographystyle{ieee_fullname}
	\bibliography{bib}
}
\end{document}